 \documentclass[final,5p,times,twocolumn]{elsarticle}

\usepackage{amssymb}
\usepackage{multirow}

\begin{document}

\begin{frontmatter}

\title{Symbiosis of an artificial neural network and models of biological neurons:\\ training and testing}

\author[inst1]{Tatyana Bogatenko}
\author[inst1]{Konstantin Sergeev}
\author[inst1]{Andrei Slepnev}
\author[inst2,inst3]{J\"urgen Kurths}
\author[inst1]{Nadezhda Semenova}\ead{semenovani@sgu.ru}

\affiliation[inst1]{organization={Saratov State University},
            addressline={83 Astrakhanskaya str.}, 
            city={Saratov},
            postcode={410012}, 
            country={Russia}}           

\affiliation[inst2]{organization={Physics Department, Humboldt University},
            addressline={15 Newtonstrasse}, 
            city={Berlin},
            postcode={12489}, 
            country={Germany}}  

\affiliation[inst3]{organization={Potsdam Institute for Climate Impact Research},
            addressline={A31 Telegrafenberg}, 
            city={Potsdam},
            postcode={14473}, 
            country={Germany}}

\begin{abstract}
In this paper we show the possibility of creating and identifying the features of an artificial neural network (ANN) which consists of mathematical models of biological neurons. The FitzHugh--Nagumo (FHN) system is used as an example of model demonstrating simplified neuron activity. First, in order to reveal how biological neurons can be embedded within an ANN, we train the ANN with nonlinear neurons to solve a a basic image recognition problem with MNIST database; and next, we describe how FHN systems can be introduced into this trained ANN. After all, we show that an ANN with FHN systems inside can be successfully trained and its accuracy becomes larger. What has been done above opens up great opportunities in terms of the direction of analog neural networks, in which artificial neurons can be replaced by biological ones.
\end{abstract}




\end{frontmatter}


\section*{Introduction}
There are two different approaches to implementing neural networks and two different definitions of neural networks. From nonlinear dynamics' perspective, a neural network is of interest for describing biological phenomena and features of interaction between neurons within a neural circuit in response to an internal or external impact. The temporal dynamics of biological neurons and the connections between them are extremely complex, so there are a large number of works describing models of different levels of complexity \cite{Hellwig2000,Sporns2000,Druckmann2014,Tewarie2018}.

On the other hand, there are artificial neural networks (ANNs). Despite having a similar name, these networks are totally different from biological neural networks in their purpose and design. Artificial neural networks have been a prospective and widely spread tool for solving many computational tasks in a variety of scientific and engineering areas branching from climate research and medicine to sociology and economy \cite{Basheer2000,Abiodun2018}.
An ANN consists of artificial neurons whose role is to generate an output signal based on a linear or nonlinear transformation of the input signal. Training an artificial neural network lies in fitting and altering the connection matrices between its neurons. In the learning process, the connection matrices are built in such a way that the network outputs the result required from it. The idea of implementing such neural networks came from biology \cite{Jain1996}. In 1940s scientists were inspired by the idea of how a neural circuit is arranged and tried to implement its simplified model in order to solve non-trivial problems that do not obtain a strictly formulated solution algorithm \cite{McCulloch1943}.

Having been first developed in the 1940s and 1950s, ANNs have undergone numerous substantial enhancements. Simple threshold neurons of the first generation, which produced binary-valued outputs, evolved into systems which use smooth activation functions, thus making it possible for the output to be real-valued. The most novel kind of an ANN is based on spiking neurons \cite{Ponulak2011,Tavanaei2019} and has received the name of a spiking neural network (SNN).

In contrast to neural networks of the previous generations, an SNN considers temporal characteristics of the information on the input. In this regard an SNN architecture makes a step closer to a plausible model of a biological neural network, although still being highly simplified \cite{GhoshDastidar2009}. Within such a network, information transmission between artificial neurons resembles that of biological neurons.

Similar to traditional ANNs, SNNs are arranged in layers, and a signal travels from an input to the output layer traversing one or more hidden layers. However, in hidden layers SNNs use spiking neurons which are described by a phenomenological model representing a spike generation process. In a biological neuron, the activity of pre-synaptic neurons affects the membrane potential of post-synaptic neurons which results in a generation of a spike when the membrane potential crosses a threshold \cite{GhoshDastidar2009,Ponulak2011}. This complex process has been described with the use of many mathematical models, with the Hodgkin-Huxley model being the first and the most famous one \cite{Hodgkin1952}. In order to find balance between computational expenses and biological realism, several other models have been proposed, e.g. the Leaky-Integrate-and-Fire model \cite{Brette2005} or the Izhikevich model \cite{Izhikevich2003}.

Based on the available knowledge from neuroscience, several methods of information encoding have been developed, e.g. rate coding or latency coding. For rate coding the rate (or the frequency) of spikes is used for information interpretation, while latency coding uses the timing of spikes. Both of these methods are special cases of a fully temporal code. In a fully temporal code a timing correlates with a certain event, for instance, a spike of a reference neuron.

Having artificial neural networks and their tasks become more and mose sophisticated, we may soon verge on some kind of a crisis \cite{Hasler2013,Gupta2015}, which consists in the fact that the tasks become so complex that the capacities of modern computers will soon not be enough to meet the growing needs. Here the bleeding-egde direction of hardware neural networks comes to the rescue \cite{Karniadakis2021}. According to this approach, neural networks are not created with a computer, but are a real device that can learn and solve tasks. The neurons themselves and the connections between them exist at the physical level, i.e. the model is not simulated on a computer, but is implemented in hardware according to its physical principles.

The main purpose of this work is to show the possibility of creating and identifying the features of a trained neural network which consists of mathematical models of biological neurons. In this research the FitzHugh-Nagumo system \cite{FitzHugh1961,Nagumo1962} in used as an example. The FHN system is a well-known simplified model of a biological neuron that demonstrates spike dynamics under certain conditions. It is often used to model simplified neural activity.

This task helps us bring artificial neural networks closer to biological ones and reveal how biological neurons can be embedded within an artificial neural network. That is, we can use the topology (the connection) from an ANN and the features of dynamics and interactions from a biological system. Thus, it is possible to approximate spike ANNs to biological ones.

The work consists of several stages. First, there is a simple neural network with artificial linear and nonlinear neurons. It is trained to solve a basic image recognition problem with handwritten digits of the MNIST database (Sect.~\ref{sec:simple_ANN}). Then, a certain number of FHN systems is introduced into the existing neural network, and the task is to identify the conditions under which the network will still function (Sect.~\ref{sec:FHN_to_ANN}). The next step is to make the task more difficult. From the beginning of this stage we use a network in which the FHN systems are implemented and attempt to train the neural network (Sect.~\ref{sec:ANN_FHN_training}).

\section{MNIST database and network topology}\label{sec:simple_ANN}
\begin{figure}[h]
\includegraphics[width=\linewidth,keepaspectratio]{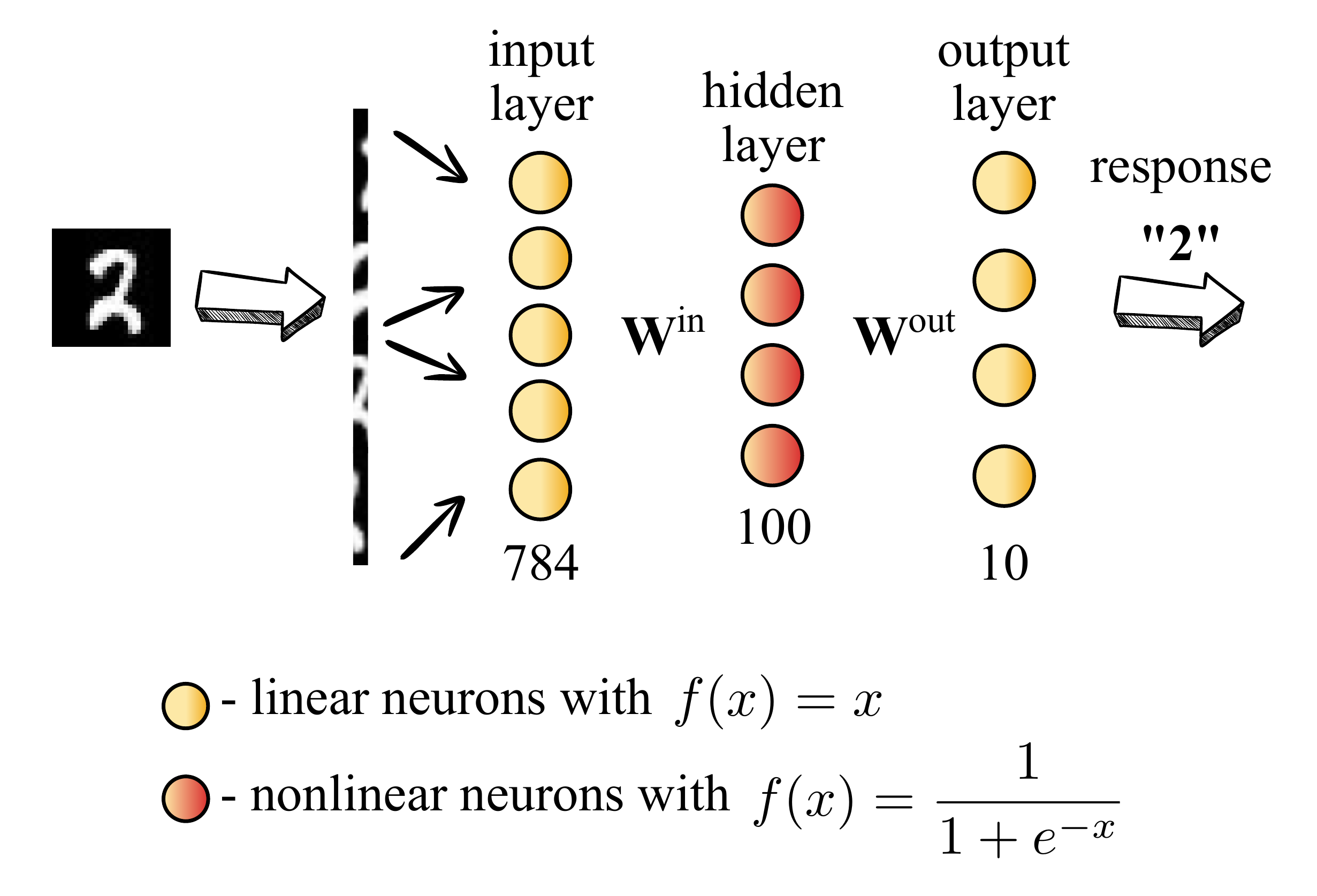}
\caption{Schematic illustration of neural network under study. The artificial neurons with linear activation function are colored in yellow, while the neurons with nonlinear activation function, which will later be replaced by FHN systems, are marked in red.}
\label{fig:ANN_scheme}
\end{figure}

At the first step a simple deep neural network with one hidden layer is being trained. The neural network is schematically shown in Fig. \ref{fig:ANN_scheme}. 

Since a task of recognising hand-written MNIST digits is being solved, an input signal of the ANN is an image of $28\times 28$ pixels. Usually such an image is tranformed into a vector $\mathbf{X}$ of size $1\times 784$ and is fed to the input layer of the ANN. This makes the first layer consist of 784 simple linear neurons with an activation function $f(x)=x$. These neurons do not transform the signal but merely pass it to the next layer. In order to do this, the vector $\mathbf{X}$ is multiplied by a corresponding matrix $\mathbf{W}^\mathrm{in}$ of size $784\times 100$. Thus, the input image is transformed into an input signal and is passed to the 100 neurons of the hidden layer. Within this layer the neurons have a sigmoid activation function: $f(x)=1/(1+e^{-x})$. However, the choice of the function does not affect the subsequent results, and the use of a hyperbolic tangent function would lead to a similar outcome. Next, the signal from the 100 hidden neurons is fed to the output layer using the connection matrix $\mathbf{W}^\mathrm{out}$. The output layer consists of 10 neurons which also use a linear activation function $f(x)=x$.

The response of the neural network is the index number of the output neuron that has the maximum output signal. This operation is also called $\mathrm{softmax}()$. Specifically, if an image with the number 2 is fed to the input of the ANN, as in Fig.~\ref{fig:ANN_scheme}, then the ANN response ``2'' will correspond to the situation when the neuron numbered $i=2$ (where $i \in[0;9]$) has the maximum output.

The MNIST database \cite{LeCun1998} was used to train the ANN. The database includes a training set (60,000 images of numbers 0--9) and a test set (10,000 images). To train the neural network, we used the Keras \cite{Chollet2015} library. This is a freely distributed API. The accuracy of the trained ANN on the training set was 99.5\%, while the accuracy of training on the test set was 97.7\%. The result of training is the connection matrices $\mathbf{W}^\mathrm{in}$ and $\mathbf{W}^\mathrm{out}$.

\section{FitzHugh--Nagumo system}\label{sec:FHN}
In order to implement FHN systems in the place of 100 artificial neurons in the hidden layer, one needs to understand the dynamics of the FHN system itself and how it should be fed with the input signal. After the input signal $\mathbf{X}$ is multiplied by the connection matrix $\mathbf{W}^\mathrm{in}$, a vector of 100 values is obtained. These values are fed to the input of 100 neurons. Now FHN systems play the role of the hidden layer neurons and each of them is described by the following equations \cite{FitzHugh1961,Nagumo1962}:
\begin{equation}\label{eq:FHN}
\begin{array}{c}
\varepsilon \dot{x} = x-\frac{x^3}{3} - y \\
\dot{y} = x+a+I(t),
\end{array}
\end{equation}
where $x$ is an activator variable, while $y$ is an inhibitor variable. This is a widely used form of the FHN system, where the parameter $\varepsilon$ is responsible for the time scale, $I(t)$ is the input signal, and $a$ is the control parameter. Depending on the value of $a$, the system demonstrates the Andronov-Hopf bifurcation: for $|a|>1$ a stable equilibrium state of the ``focus'' type is observed (excitable mode); if $|a|<1$, the mode is called oscillatory, and the system exhibits periodic spike dynamics.
 
A description of the system (\ref{eq:FHN}) from the perspective of current and voltage is also common. Then the variable $x$ is a voltage-like membrane potential with cubic nonlinearity that allows regenerative self-excitation via a positive feedback. The variable $y$ is called the recovery variable with linear dynamics that provides a slower negative feedback. The parameter $I$ corresponds to a stimulus current. A positive current corresponds to a current directed from the outside of the cell membrane to the inside.

\begin{figure}[h]
\includegraphics[width=\linewidth,keepaspectratio]{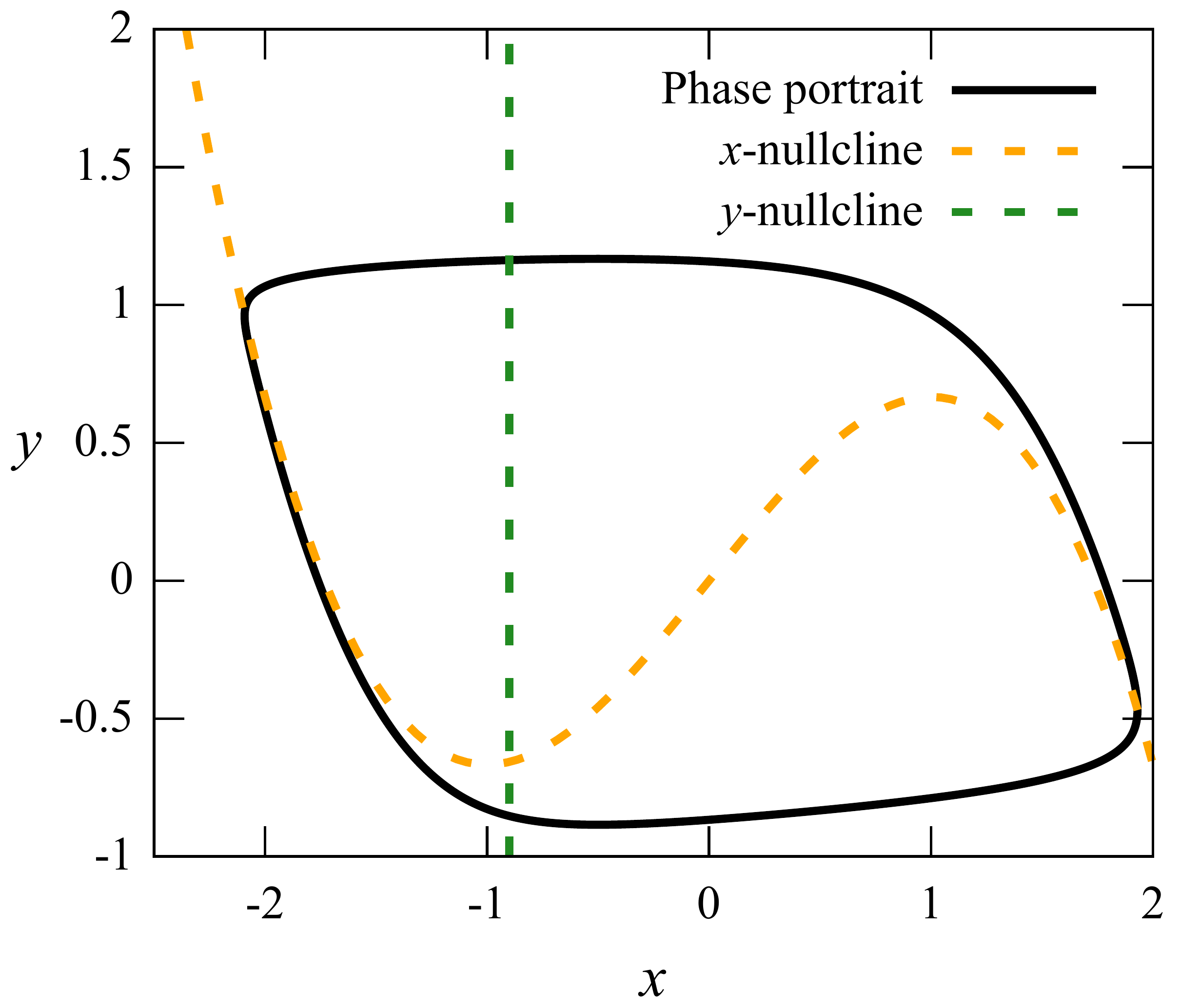}
\caption{Phase portrait of FHN system (\ref{eq:FHN}) with corresponding $\dot{x}=0$ (orange) and $\dot{y}=0$ nullclines (green).}
\label{fig:FHN_portrait}
\end{figure}

Figure ~\ref{fig:FHN_portrait} shows the phase plane of the system (\ref{eq:FHN}). Also, the corresponding activator $\dot{x}=0$ and inhibitor $\dot{y}=0$ nullclines are depicted. The activator nullcine corresponds to the $y=x-x^3/3$ line (Fig.~\ref{fig:FHN_portrait}, the orange line), while the inhibitor nullcline corresponds to the $x=-a$ line when there is no input signal (Fig.~\ref{fig:FHN_portrait}, the green line). For $a=1$ the nullclines intersect at $x_0=-1$, $y_0=-2/3$. This point is a stable equilibrium state for $a>1$.

If the input signal does not change in time but introduces an additional constant component $I(t)=I=\mathrm{const}$ into the second equation, the sum $a+I$ allows one to influence the position of the vertical nullcline of the system (see Fig. ~\ref{fig:FHN_portrait}, green dash line). Thus, due to the input signal $I$, the system can establish either an oscillatory or excitable mode. For $a=0$ the values $|I|<1$ will correspond to the oscillatory mode, and $|I|>1$ will establish the excitable one.

\section{Implementing the FHN systems into the trained ANN}\label{sec:FHN_to_ANN}
Since the neural network was initially trained in such a way that the hidden layer neurons have a ``sigmoid'' type activation function, they accept an input signal of the range $(-\infty;+\infty)$, and return an output signal of $(0;1)$. In this case, the product of the input image vector $\mathbf{X}$ and the connection matrix $\mathbf{W}^\mathrm{in}$ may contain such large numbers that they will not be commensurate with the scale of the variables $(x ,y)$ of the FHN system.

In order for the product $\mathbf{X}\cdot\mathbf{W}^\mathrm{in}$ to be further used as an input signal of the FHN system, we propose to introduce the following normalization:
\begin{equation}\label{eq:I_norm}
I = \gamma\cdot\tanh(\mathbf{X}\cdot\mathbf{W}^\mathrm{in}).
\end{equation}
Then, no matter how large the values of the matrix $\mathbf{W}^\mathrm{in}$ are, after applying the hyperbolic tangent the range of values is transformed into $(-1;1)$. The $\gamma$ multiplier allows you to set the range of the values more precisely.

The output signal of the system is defined as follows:
\begin{equation}\label{eq:FHN_Y}
Y=\mathrm{softmax}(\vec{x}\cdot \mathbf{W}^\mathrm{out}),
\end{equation}
It is also computed with the use of the $\mathrm{softmax}()$ function as earlier, but now it is a function of the product of the $\mathbf{W}^\mathrm{out}$ matrix and the variable vector $x_i$ of the 100 FHN systems. This makes the ANN response to be the index number of the output layer neuron which has the maximum output signal.

\begin{figure}[h]
\includegraphics[width=\linewidth,keepaspectratio]{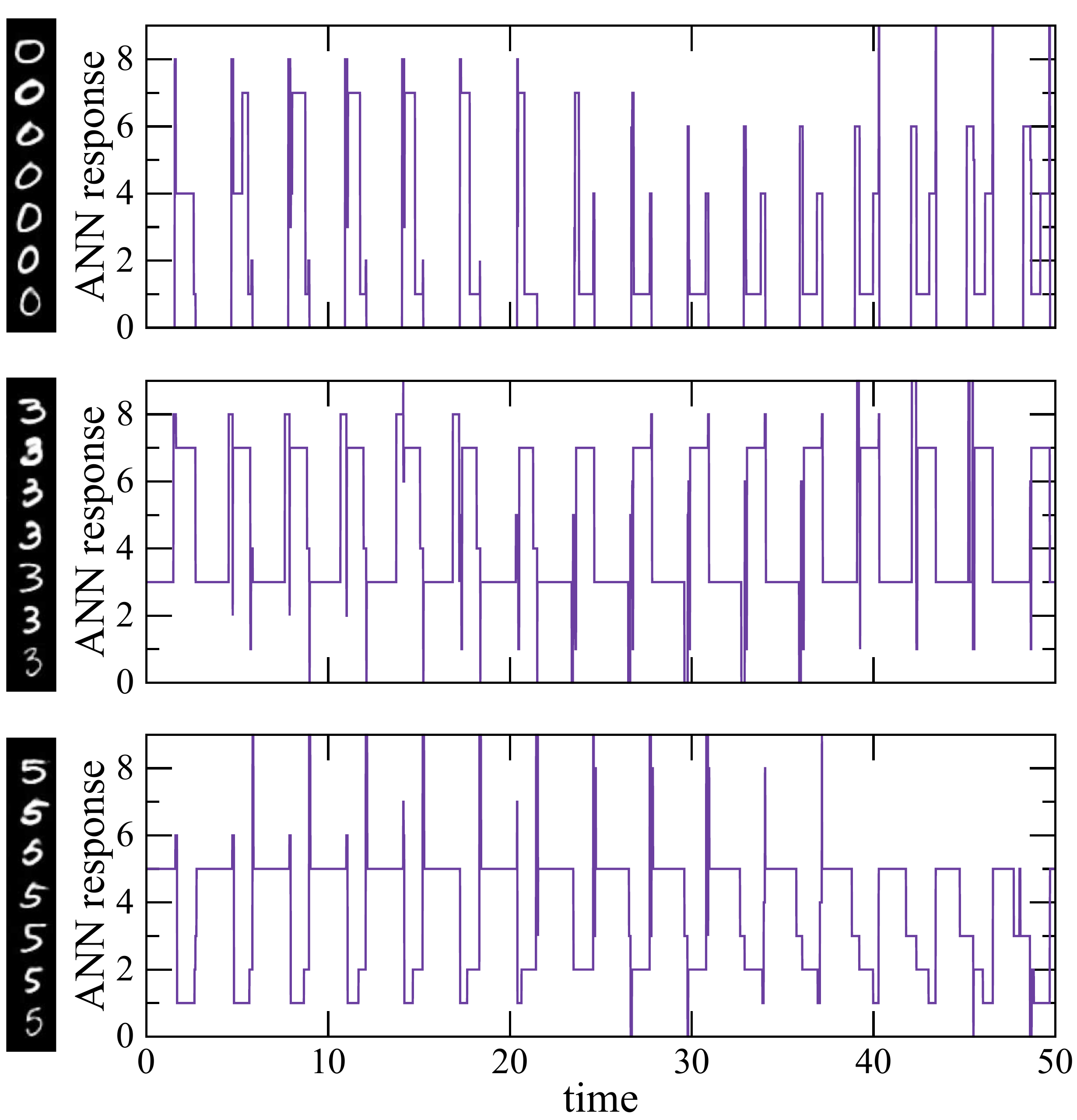}
\caption{Temporal evolution of the output of ANN with implemented FHN systems for three different kinds of input image: digit ``0'' (top panel), ``3'' (middle) and ``5'' (bottom).}
\label{fig:ANN_FHN_realization}
\end{figure}

Figure ~\ref{fig:ANN_FHN_realization} shows temporal dependencies of the neural network response for three different input images containing the numbers 0, 3 and 5. Strictly speaking, Fig.~\ref{fig:ANN_FHN_realization} does not show an immediate output of the neural network. There is a transient time of 1000 dimensionless units $T^\mathrm{trans}=1000$ which is discarded in order not to consider the regime establishment process. Now there is a problem of the result interpretation. Since FHN systems are spike systems and may show an oscillatory mode, the ANN response may also oscillate, and at some moments ``wrong'' neurons can be activated. In order to interpret the response of the ANN correctly, one can choose the answer that takes the most time of the entire control record of the ANN output signal $T=100$. For example, in the Fig.~\ref{fig:ANN_FHN_realization}, the longest response time is ``0'' for the top record, ``3'' for the middle one, and ``5'' for the bottom one.

Identical results were obtained for other digits. The table \ref{tab:ANN_FHN} shows the accuracy calculated for the digits 0--9 from the training and testing sets for the parameter $\gamma=-0.5$. As can be seen from the table, the average accuracy on the training set was 73.4\%, and the average accuracy on the test set was 73.3\%.

\begin{table}
\begin{center}
\begin{tabular}{| c | c | c | c | c | c | c | c |} 
\hline
\multirow{2}{2em}{Digit} & \multicolumn{2}{|c|}{ $\gamma=-0.5$ } & \multicolumn{2}{|c|}{ $\gamma=-1$ } & \multicolumn{2}{|c|}{ $\gamma=0.5$ }  \\
  & Tr. & Test. & Tr. & Test. & Tr. & Test. \\
 \hline
 \hline
0  &  99.0  &  98.6  &  88.9  &  99.2  &  6.7   &  6.4  \\
1  &  1.4   &  1.5   &  16.3  &  16.4  &  0.0   &  0    \\
2  &  83.3  &  82.6  &  91.8  &  90.3  &  5.6   &  5.6  \\
3  &  84.6  &  85.1  &  92.7  &  91.9  &  3.2   &  3.7  \\
4  &  90.4  &  88.0  &  98.0  &  94.3  &  1.0   &  0.6  \\
5  &  29.6  &  31.9  &  52.0  &  54.9  &  0.3   &  1.1  \\
6  &  96.9  &  96.2  &  99.9  &  98.0  &  8.6   &  9.4  \\
7  &  56.5  &  57.5  &  80.4  &  78.6  &  0.9   &  0.8  \\
8  &  99.8  &  99.5  &  100.0 &  99.0  &  6.5   &  6.0  \\
9  &  92.3  &  91.9  &  98.9  &  96.6  &  1.36  &  1.98  \\
 \hline
 \hline
Average & 73.4 & 73.3 & 83.0 & 82.9 & 3.4 & 3.6 \\
 \hline
 \hline
\end{tabular}
\end{center}
\caption{Accuracies of ANN with implemented FHN systems applied to training (Tr.) and testing (Test.) datasets of each digit type. There are three ANNs with $\gamma=-0.5$, $\gamma=-1$ and $\gamma=0.5$.}
\label{tab:ANN_FHN}
\end{table}

\begin{figure}[h]
\includegraphics[width=\linewidth,keepaspectratio]{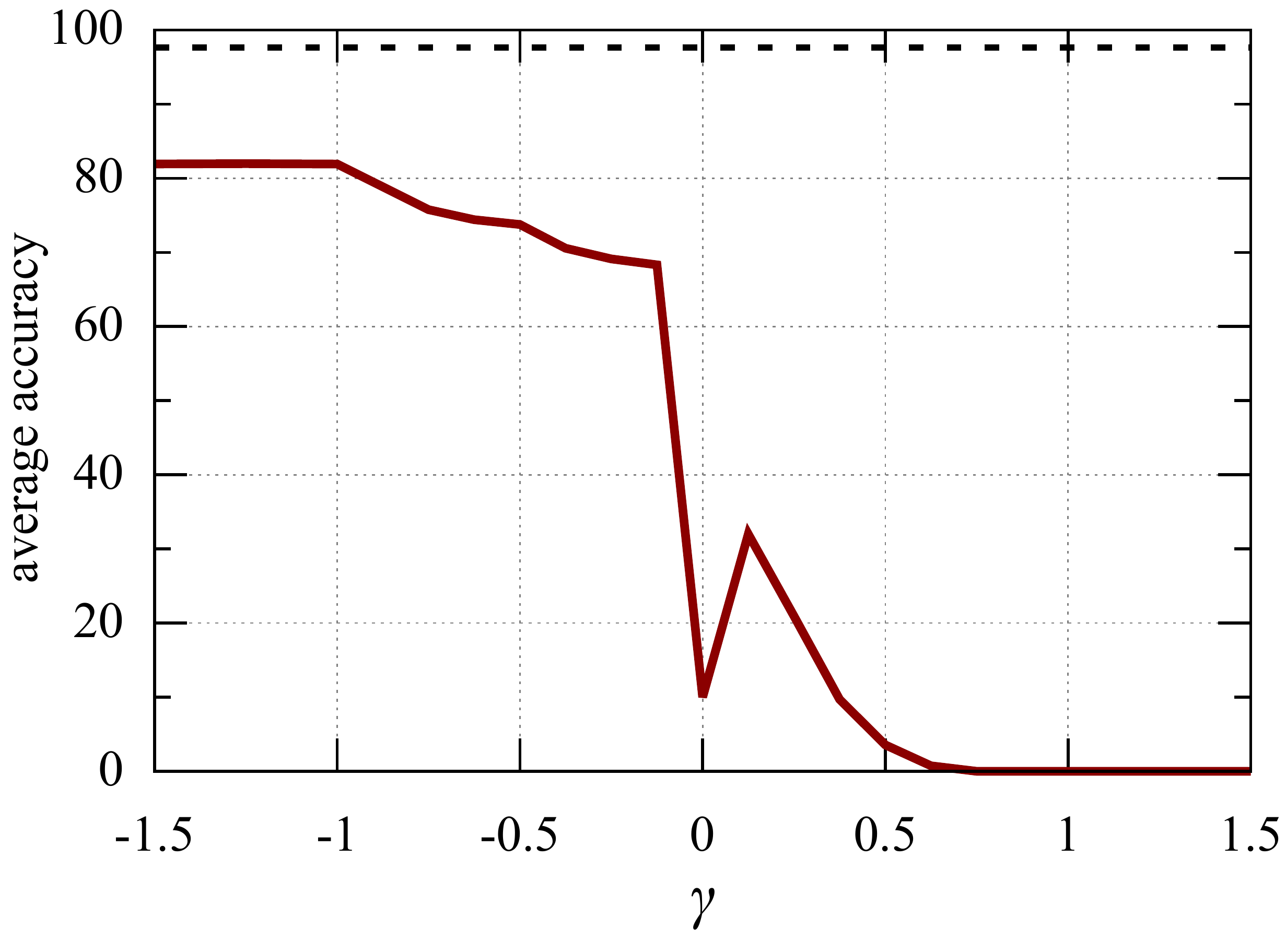}
\caption{Average testing accuracy of ANN with implemented FHN systems depending on the parameter $\gamma$.}
\label{fig:ANN_FHN_gamma}
\end{figure}

Average accuracies were also calculated for other values of $\gamma$. Figure ~\ref{fig:ANN_FHN_gamma} shows the dependence of the average accuracy on $\gamma$ for the test set. As can be seen from the picture, the lowest accuracy can be obtained if $\gamma>0$ and if $\gamma$ is close to zero. Also, for $-1<\gamma<0$, the accuracy increases with the decrease of $\gamma$ and saturates when $\gamma<-1$.

\section{ANN training}\label{sec:ANN_FHN_training}
In order to speed up the processes of training and testing, not all the images from the training and test sets were used . 10000 examples from the training set and 1000 examples of the test set with equal amount of examples of the same digit were used.

The parameters of the FHN systems remained the same. Since ANN training is associated with a large number of runs of input images and connection matrices, in order to speed up this process, the settling time was reduced to $T^\mathrm{trans}=200$, but the control time $T=100$ remained the same. This did not affect the accuracy when repeating the previously described steps.

\begin{figure}[h]
\includegraphics[width=\linewidth,keepaspectratio]{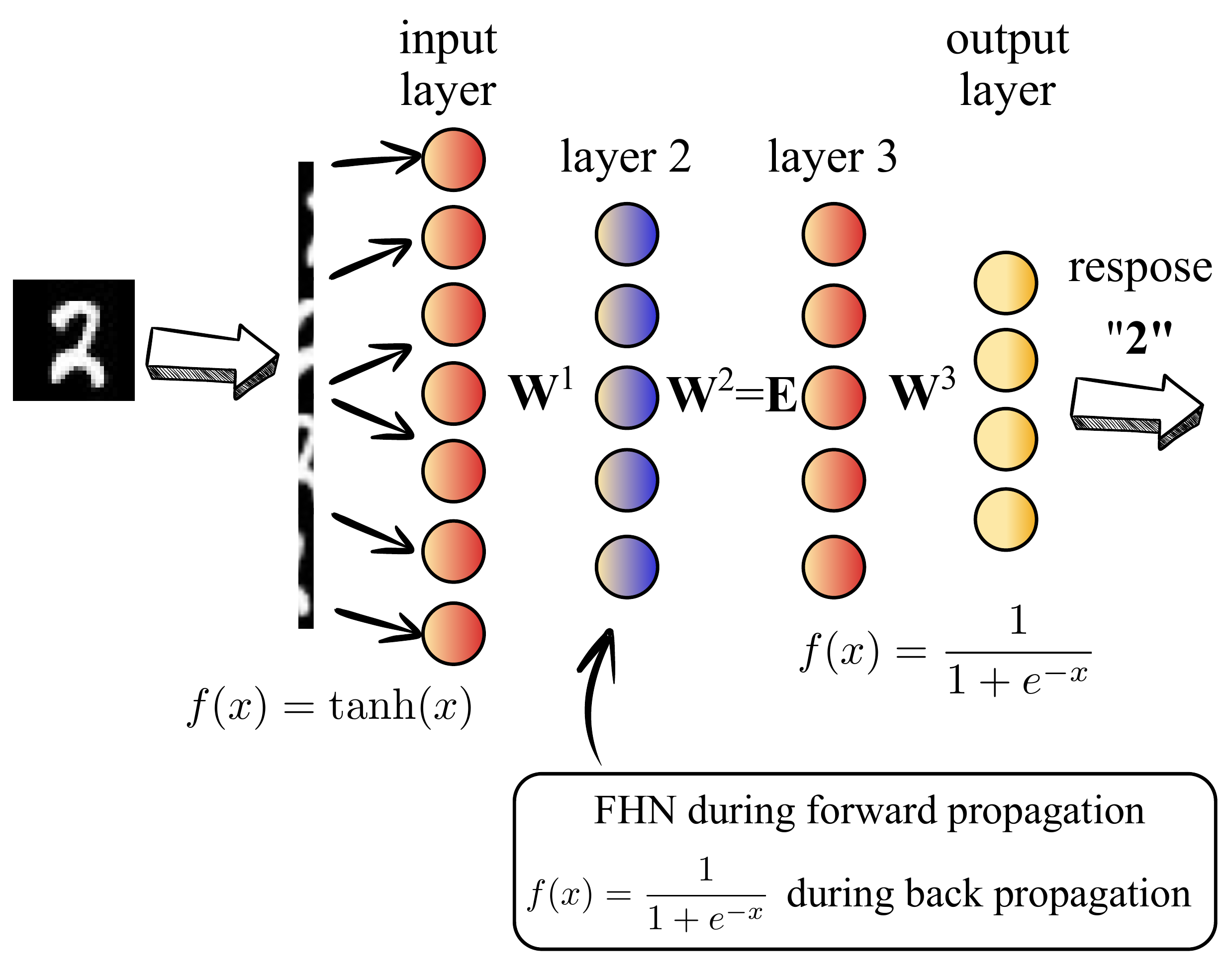}
\caption{Schematic illustration of trainable ANN with FHN systems. The artificial neurons with linear activation function are colored in yellow, while the neurons with nonlinear activation function are marked in red. Violet color corresponds to FHN systems.}
\label{fig:ANN_FHN_scheme}
\end{figure}

There were some difficulties in the process of training the ANN. A large number of network topologies and several nonlinear activation functions were considered, the number of layers was also being changed. In the end, we came to the conclusion that the optimal network topology looks like the one presented in the Fig.~\ref{fig:ANN_FHN_scheme}. 784 pixels are fed to the ANN input, so the input layer still consists of 784 neurons. In the previous section, it was shown that before applying a signal to the FHN input, it must first be renormalized using a hyperbolic tangent, so we added this processing step to the first layer in the new ANN. As a result, the number of neurons in the first layer remains the same, but their activation function becomes a hyperbolic tangent. The second layer is 100 FHN systems. The input layer is connected to the second layer by a $\mathbf{W}^1$ matrix $784\times 100$.

The third layer was necessary to simplify the processes of learning and calculating derivatives. It contains 100 artificial neurons with a ``sigmoid'' activation function. Layers 2 and 3 are interconnected using the idenity connection matrix $\mathbf{W}^2=E$, which is fixed and does not change during the learning process, i.e. neurons of layers 2 and 3 are connected one-to-one throughout the training.

The output layer contains 10 linear neurons with the function $\mathrm{softmax}()$. The output layer is connected with the previous one with the connection matrix $\mathbf{W}^3$.

The interpretation of the output signal of the obtained ANN was the same as in the previous section. An ANN's response over time $T=100$ was the index number of the neuron which produced the largest output signal for the longest time.

\begin{figure}[h]
\includegraphics[width=\linewidth,keepaspectratio]{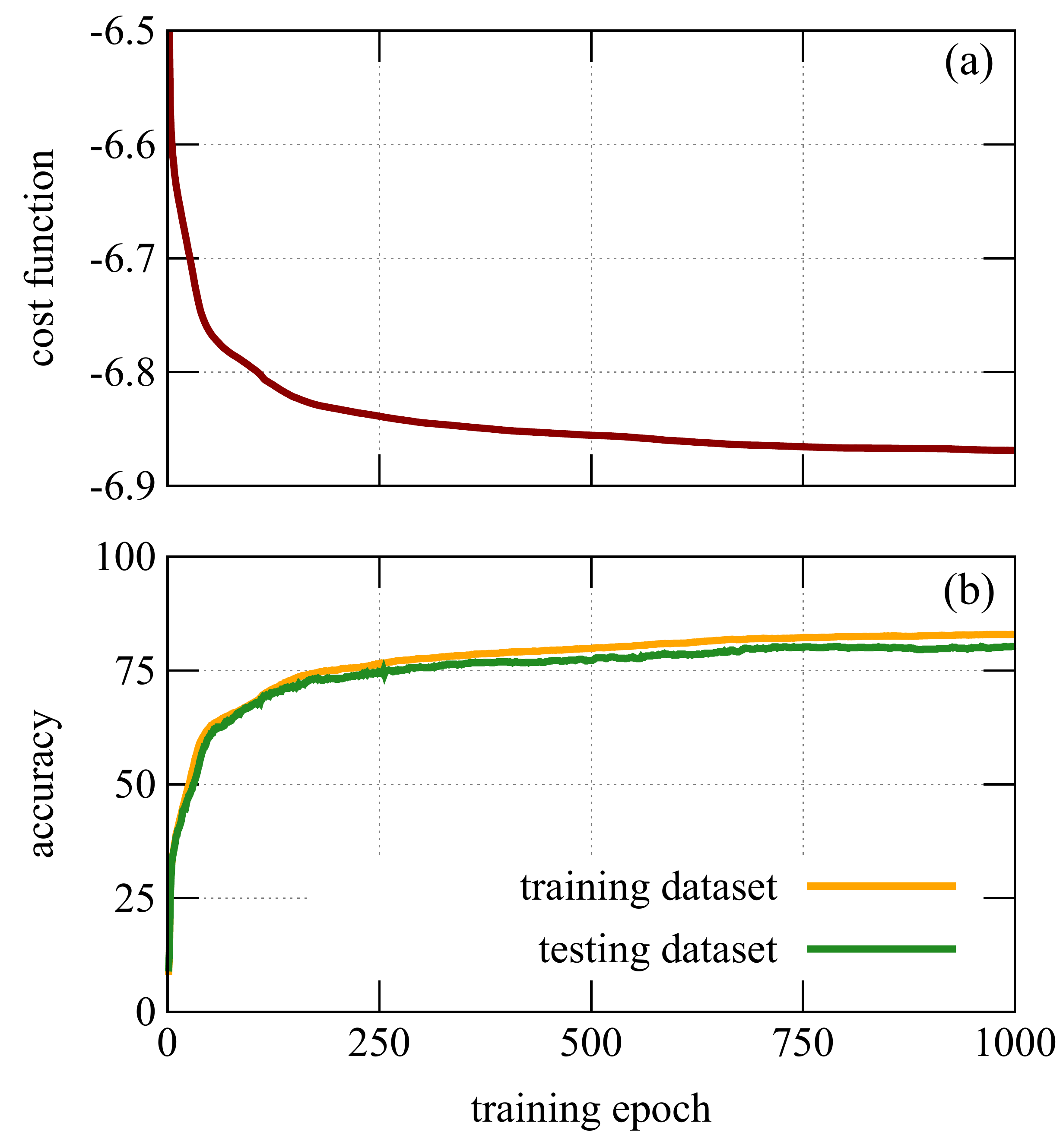}
\caption{Training process of ANN with FHN systems inside illustrated by cost function (a) and accuracies (b) on training and testing datasets depending on the training epoch.}
\label{fig:ANN_FHN_training}
\end{figure}

The training was carried out using the backpropagation method of linear regression. Here the following trick is applied. During the forward propagation, FHN systems were used as layer 2 neurons, and during back propagation, they were replaced by conventional artificial neurons with a ``sigmoid'' type activation function $f(x)=1/(1+e^{-x})$ for the correct calculation of the derivative. In Fig.~\ref{fig:ANN_FHN_scheme} these neurons are represented in blue. In Fig.~\ref{fig:ANN_FHN_training} the cost function (a) and accuracy on the training and test sets depending on the training epoch (b) illustrate the training process.

The corresponding accuracies on the training and test sets by digits are given in Table \ref{tab:ANN_FHN_trained} for both the truncated sets (for which Fig.~\ref{fig:ANN_FHN_training} was built) and the full MNIST sets (for which we have already applied the trained network). The above results were obtained for the value $\gamma=-0.5$. The average accuracy of the resulting network is about 80\%. In comparison of this value with the accuracy of the trained ANN with FHN systems embedded in it, this $\gamma$ value corresponds to an accuracy of about 73\%. Thus, after training with the proposed technique, it was possible to increase the accuracy of the neural network.

\begin{table}
\begin{center}
\begin{tabular}{| c | c | c | c | c |} 
\hline
\multirow{2}{2em}{Digit} & Tr. set & Test. set & Tr. set & Test. set \\
 & (10000) & (1000) & (60000) & (10000) \\
\hline
\hline

0 & 97.1 & 96.5 & 95.9 & 95.6 \\
1 & 97.5 & 97.6 & 96.6 & 97.8 \\
2 & 89.2 & 87.1 & 86.4 & 86.6 \\
3 & 87.1 & 85.1 & 84.2 & 86.8 \\
4 & 91.9 & 86.4 & 88.6 & 89.1 \\
5 & 85.9 & 86.2 & 81.0 & 81.7 \\
6 & 95.2 & 86.2 & 92.2 & 91.1 \\
7 & 94.4 & 86.9 & 92.2 & 90.5 \\
8 &  0.1 &  0   &  0.1 &  0.1 \\
9 & 83.7 & 80.9 & 79.5 & 79.9 \\
 \hline
 \hline
Average & 82.2 & 79.3 & 79.7 & 79.9 \\
 \hline
 \hline
\end{tabular}
\end{center}
\caption{Accuracies of trained ANN with FHN systems applied to training (Tr.) and testing (Test.) datasets of each digit type. Parameter $\gamma$ is set $-0.5$.}
\label{tab:ANN_FHN_trained}
\end{table}

The distribution of accuracy for different digits is of particular interest (see Table \ref{tab:ANN_FHN_trained} and Fig.~\ref{fig:ANN_FHN_acc_digits} in Appendix). The resulting ANN does not work well with the number 8. If it is not considered, the overall accuracy is about 90\%.

\section*{Conclusion}
We managed to find ways to introduce FitzHugh-Nagumo systems into artificial neural networks, which made it possible to combine the neural network topology with the peculiarities of the FitzHugh-Nagumo spike dynamics. We were also able to find the conditions under which the resulting neural network demonstrates good accuracy. 

The signal from the first layer is multiplied by the corresponding connection matrix and then it is sent to each of 100 FHN systems in ANNs' hidden layer according to Eq.~(\ref{eq:I_norm}). The multiplier $\gamma$ allows to control the amplitude of FHN input signal more precisely. We show here that only negative $\gamma$ values lead to appropriate accuracy.

In addition, we proposed a method for training the ANN with introduced FHN systems. It is schematically shown in Fig.~\ref{fig:ANN_FHN_training}. The resulting neural network, which increases the accuracy by $\approx 12 \%$.

\section*{Data availability}
The data that support the findings of this study are available from the corresponding author upon reasonable request.

\section*{Acknowledgements}
This work is partially supported by the Russian President scholarship SP-749.2022.5.

\appendix
\section{Testing accuracy of trained ANN according to digit type}
\label{sec:appendix}

\begin{figure}[h]
\includegraphics[width=\linewidth,keepaspectratio]{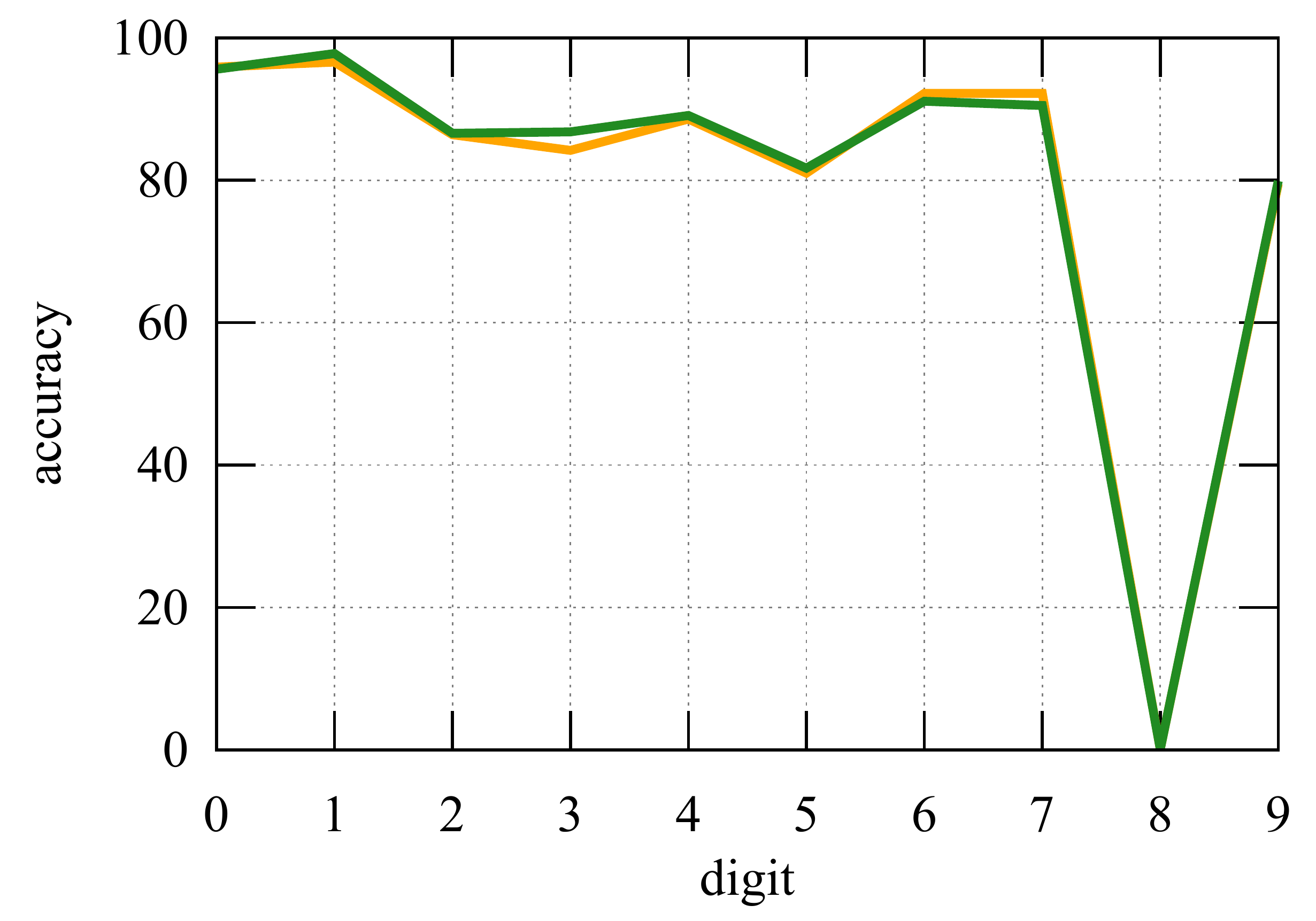}
\caption{Testing accuracy of the data from Table~\ref{tab:ANN_FHN_trained} grouped by the digit type.}
\label{fig:ANN_FHN_acc_digits}
\end{figure}


\begin{thebibliography}{10}
\expandafter\ifx\csname url\endcsname\relax
  \def\url#1{\texttt{#1}}\fi
\expandafter\ifx\csname urlprefix\endcsname\relax\def\urlprefix{URL }\fi
\expandafter\ifx\csname href\endcsname\relax
  \def\href#1#2{#2} \def\path#1{#1}\fi

\bibitem{Hellwig2000}
B.~Hellwig, A quantitative analysis of the local connectivity between pyramidal
  neurons in layers 2/3 of the rat visual cortex, Biological Cybernetics 82~(2)
  (2000) 111--121.
\newblock \href {https://doi.org/10.1007/PL00007964}
  {\path{doi:10.1007/PL00007964}}.

\bibitem{Sporns2000}
O.~Sporns, G.~Tononi, G.~Edelman, Connectivity and complexity: the relationship
  between neuroanatomy and brain dynamics, Neural Networks 13~(8) (2000)
  909--922.
\newblock \href {https://doi.org/10.1016/S0893-6080(00)00053-8}
  {\path{doi:10.1016/S0893-6080(00)00053-8}}.

\bibitem{Druckmann2014}
S.~Druckmann, L.~Feng, B.~Lee, C.~Yook, T.~Zhao, J.~Magee, J.~Kim, Structured
  synaptic connectivity between hippocampal regions, Neuron 81~(3) (2014)
  629--640.
\newblock \href {https://doi.org/10.1016/j.neuron.2013.11.026}
  {\path{doi:10.1016/j.neuron.2013.11.026}}.

\bibitem{Tewarie2018}
P.~Tewarie, B.~A.~E. Hunt, G.~C. O’Neill, A.~Byrne, K.~Aquino, M.~Bauer,
  K.~J. Mullinger, S.~Coombes, M.~J. Brookes, {Relationships Between Neuronal
  Oscillatory Amplitude and Dynamic Functional Connectivity}, Cerebral Cortex
  29~(6) (2018) 2668--2681.
\newblock \href {https://doi.org/10.1093/cercor/bhy136}
  {\path{doi:10.1093/cercor/bhy136}}.

\bibitem{Basheer2000}
I.~Basheer, M.~Hajmeer, Artificial neural networks: fundamentals, computing,
  design, and application, Journal of Microbiological Methods 43~(1) (2000)
  3--31, neural Computting in Micrbiology.
\newblock \href {https://doi.org/10.1016/S0167-7012(00)00201-3}
  {\path{doi:10.1016/S0167-7012(00)00201-3}}.

\bibitem{Abiodun2018}
O.~I. Abiodun, A.~Jantan, A.~E. Omolara, K.~V. Dada, N.~A. Mohamed, H.~Arshad,
  State-of-the-art in artificial neural network applications: A survey, Heliyon
  4~(11) (2018) e00938.
\newblock \href {https://doi.org/10.1016/j.heliyon.2018.e00938}
  {\path{doi:10.1016/j.heliyon.2018.e00938}}.

\bibitem{Jain1996}
A.~Jain, J.~Mao, K.~Mohiuddin, Artificial neural networks: a tutorial, Computer
  29~(3) (1996) 31--44.
\newblock \href {https://doi.org/10.1109/2.485891}
  {\path{doi:10.1109/2.485891}}.

\bibitem{McCulloch1943}
W.~S. McCulloch, W.~Pitts, A logical calculus of the ideas immanent in nervous
  activity, The bulletin of mathematical biophysics 5~(4) (1943) 115--133.
\newblock \href {https://doi.org/10.1007/BF02478259}
  {\path{doi:10.1007/BF02478259}}.

\bibitem{Ponulak2011}
F.~Ponulak, A.~Kasinski,
  \href{http://europepmc.org/abstract/MED/22237491}{Introduction to spiking
  neural networks: Information processing, learning and applications}, Acta
  neurobiologiae experimentalis 71~(4) (2011) 409—433.
\newline\urlprefix\url{http://europepmc.org/abstract/MED/22237491}

\bibitem{Tavanaei2019}
A.~Tavanaei, M.~Ghodrati, S.~R. Kheradpisheh, T.~Masquelier, A.~Maida, Deep
  learning in spiking neural networks, Neural Networks 111 (2019) 47--63.
\newblock \href {https://doi.org/10.1016/j.neunet.2018.12.002}
  {\path{doi:10.1016/j.neunet.2018.12.002}}.

\bibitem{GhoshDastidar2009}
S.~Ghosh-Dastidar, H.~Adeli, Spiking neural networks, International Journal of
  Neural Systems 19~(04) (2009) 295--308, pMID: 19731402.
\newblock \href {https://doi.org/10.1142/S0129065709002002}
  {\path{doi:10.1142/S0129065709002002}}.

\bibitem{Hodgkin1952}
A.~L. Hodgkin, A.~F. Huxley, A quantitative description of membrane current and
  its application to conduction and excitation in nerve, The Journal of
  physiology 117~(4) (1952) 500.
\newblock \href {https://doi.org/10.1113/jphysiol.1952.sp004764}
  {\path{doi:10.1113/jphysiol.1952.sp004764}}.

\bibitem{Brette2005}
R.~Brette, W.~Gerstner, Adaptive exponential integrate-and-fire model as an
  effective description of neuronal activity, Journal of Neurophysiology 94~(5)
  (2005) 3637--3642, pMID: 16014787.
\newblock \href {https://doi.org/10.1152/jn.00686.2005}
  {\path{doi:10.1152/jn.00686.2005}}.

\bibitem{Izhikevich2003}
E.~Izhikevich, Simple model of spiking neurons, IEEE Transactions on Neural
  Networks 14~(6) (2003) 1569--1572.
\newblock \href {https://doi.org/10.1109/TNN.2003.820440}
  {\path{doi:10.1109/TNN.2003.820440}}.

\bibitem{Hasler2013}
J.~Hasler, H.~Marr, Finding a roadmap to achieve large neuromorphic hardware
  systems, Frontiers in Neuroscience 7 (2013).
\newblock \href {https://doi.org/10.3389/fnins.2013.00118}
  {\path{doi:10.3389/fnins.2013.00118}}.

\bibitem{Gupta2015}
S.~Gupta, A.~Agrawal, K.~Gopalakrishnan, P.~Narayanan,
  \href{https://proceedings.mlr.press/v37/gupta15.html}{Deep learning with
  limited numerical precision}, in: F.~Bach, D.~Blei (Eds.), Proceedings of the
  32nd International Conference on Machine Learning, Vol.~37 of Proceedings of
  Machine Learning Research, PMLR, Lille, France, 2015, pp. 1737--1746.
\newline\urlprefix\url{https://proceedings.mlr.press/v37/gupta15.html}

\bibitem{Karniadakis2021}
G.~E. Karniadakis, I.~G. Kevrekidis, L.~Lu, P.~Perdikaris, S.~Wang, L.~Yang,
  Physics-informed machine learning, Nature Reviews Physics 3~(6) (2021)
  422--440.
\newblock \href {https://doi.org/10.1038/s42254-021-00314-5}
  {\path{doi:10.1038/s42254-021-00314-5}}.

\bibitem{FitzHugh1961}
R.~FitzHugh, Impulses and physiological states in theoretical models of nerve
  membrane, Biophysical Journal 1~(6) (1961) 445--466.
\newblock \href {https://doi.org/10.1016/S0006-3495(61)86902-6}
  {\path{doi:10.1016/S0006-3495(61)86902-6}}.

\bibitem{Nagumo1962}
J.~Nagumo, S.~Arimoto, S.~Yoshizawa, An active pulse transmission line
  simulating nerve axon, Proceedings of the IRE 50~(10) (1962) 2061--2070.
\newblock \href {https://doi.org/10.1109/JRPROC.1962.288235}
  {\path{doi:10.1109/JRPROC.1962.288235}}.

\bibitem{LeCun1998}
Y.~LeCun, \href{http://yann. lecun. com/exdb/mnist/}{The mnist database of
  handwritten digits} (1998).
\newline\urlprefix\url{http://yann. lecun. com/exdb/mnist/}

\bibitem{Chollet2015}
F.~Chollet, et~al.
\newblock \href{https://github.com/fchollet/keras}{Keras} [online] (2015).

\end{thebibliography}

\end{document}